\documentclass[accepted]{uai2021} 

\usepackage[american]{babel}

\usepackage{natbib} 
\usepackage{mathtools} 
\usepackage{booktabs} 
\usepackage{tikz} 

\usepackage{algorithm, algorithmic}
\usepackage{amsmath}
\usepackage{amsfonts}
\usepackage{booktabs}
\usepackage{multirow}
\usepackage{subcaption}

\title{Natural Language Adversarial Defense through Synonym Encoding}

\author[ ]{Xiaosen Wang}
\author[ ]{Hao Jin}
\author[ ]{Yichen Yang}
\author[ ]{Kun He\thanks{Corresponding author.}}

\affil[ ]{%
    School of Computer Science and Technology\\
    Huazhong University of Science and Technology\\
    Wuhan\\
    China 
}

\affil[ ]{
\texttt{ \{xiaosen,mailtojinhao,yangyc,brooklet60\}@hust.edu.cn}
}

\begin{document}
\maketitle

\begin{abstract}
  In the area of natural language processing, deep learning models are recently known to be vulnerable to various types of adversarial perturbations, but relatively few works are done on the defense side. Especially, there exists few effective defense method against the successful synonym substitution based attacks that preserve the syntactic structure and semantic information of the original text while fooling the deep learning models. We contribute in this direction and propose a novel adversarial defense method called \textit{Synonym Encoding Method} (SEM). Specifically, SEM inserts an encoder before the input layer of the target model to map each cluster of synonyms to a unique encoding and trains the model to eliminate possible adversarial perturbations without modifying the network architecture or adding extra data. Extensive experiments demonstrate that SEM can effectively defend the current synonym substitution based attacks and block the transferability of adversarial examples. SEM is also easy and efficient to scale to large models and big datasets.
\end{abstract}

\section{Introduction}\label{sec:intro}
Deep Neural Networks (DNNs) have made great success in various machine learning tasks. However, recent studies have found that DNNs are often vulnerable to \textit{adversarial examples}, in which the original examples are modified imperceptibly to humans but could mislead deep learning models. More seriously, the adversaries are found not only in computer vision tasks~\citep{Szegedy2014Intriguing} but even in Natural Language Processing (NLP) tasks~\citep{papernot2016crafting}, raising security and safety concerns. For instance, spammers can evade the spam filtering system with crafted adversarial examples of spam emails while preserving the intended meaning. 

In contrast to the fact that numerous methods have been proposed in the area of computer vision for adversarial attacks \citep{Goodfellow2015Explaining, Carlini2017CW, Athalye2018Obfuscated,Dong2018MIM,guo2019low,wang2021enhancing} and defenses \citep{Goodfellow2015Explaining, Chuan2018Countering, Biao2019improving}, there are relatively few works done in the area of NLP. Works on text adversaries just emerge in recent years, and most of them are inspired by methods proposed for images~\citep{NLPAttackSurvey}. 
However, existing adversarial learning methods for images could not be directly applied to texts due to the discrete property of texts in nature. Furthermore, if we want a crafted text perturbation to be barely perceptible to humans, it should maintain the lexical and grammatical correctness and preserve the original semantic information, making it harder to craft the textual adversarial examples.

Current adversarial attacks in NLP roughly invoke one or several of the following methods: modifying the characters within a word \citep{liang2017deep, ebrahimi2017hotflip, li2018textbugger}, adding or removing words \citep{liang2017deep, samanta2017crafting}, replacing words based on embedding perturbations \citep{papernot2016crafting,gong2018adversarial}, substituting words with synonyms \citep{samanta2017crafting,alzantot2018generating,Shuhuai2019wordsaliency}, and crafting paraphrases for the entire sentence \citep{iyyer2018adversarial,ribeiro2018semantically}. However, perturbations on characters or words that destroy the syntax can be easily detected and defended by the spelling or syntax check~\citep{Rodriguez2018Shielding,pruthi2019combating}. Moreover, both paraphrasing and word replacement determined by the embedding perturbations usually face the challenge of ensuring the preservation of the original semantics. As synonym substitution aims to satisfy the lexical, grammatical and semantic constraints, it is much harder to be detected by automatic spelling or syntax check as well as human investigation, and hence synonym substitution is more efficacious for textual adversarial attacks. 

On the defense side for synonym substitution based attacks, \cite{alzantot2018generating} and \cite{Shuhuai2019wordsaliency} incorporate the perturbed examples during the training in an attempt to improve the model robustness, but witness an insufficient amount of adversarial examples for the adversarial training due to the low efficiency of adversary generation. Another line of work~\citep{jia2019certified,huang2019achieving} is towards certified robustness and based on Interval Bound Propagation (IBP)~\citep{Gowal2019IBP}. However, such defense methods are hard to scale to big datasets and large neural networks for their high complexity and also result in lower accuracy on benign data due to the loose upper bounds. 

In this work, we propose a novel defense method against synonym substitution based attacks. Specifically, we postulate that a generalization that is not strong enough usually results in different classification results for the neighbors \(\{x'|x'\in V_\epsilon (x)\}\) of a benign example \(x\) in the data manifold. Based on this hypothesis, we propose a new defense paradigm called \textit{Synonym Encoding Method} (SEM) that encodes each cluster of synonyms to a unique encoding so as to force all the neighbors of an input text \(x\) to share the same code of \(x\). Specifically, we first cluster the synonyms according to the \textit{Euclidean distance} in the embedding space to construct the encoder. Then, we insert the encoder before the input layer of a deep learning model without modifying its architecture and train the model with such an encoder on the original dataset to effectively defend the adversarial attacks in the context of text classification.

The proposed method is simple, efficient and highly scalable. Experiments on three popular datasets demonstrate that SEM can effectively defend synonym substitution based adversarial attacks and block the transferability of adversarial examples in the context of text classification. Also, SEM maintains computational efficiency and is easy to scale to large neural networks and big datasets without modifying the network architecture or using extra data. Meanwhile, SEM achieves almost the same accuracy on benign data as the original model does, and the accuracy is higher than that of the certified defense method IBP. 

\section{Background} \label{sec:bg}
Let \(\mathcal{W}\) denote the dictionary containing all the legal words. Let \(x=\langle w_1, \dots, w_i, \dots, w_n \rangle\) denote an input text, \(\mathcal{C}\) the corpus that contains all the possible input texts, and \(\mathcal{Y} \in \mathbb{N}^K\) the output space where \(K\) is the dimension of \(\mathcal{Y}\). The classifier \(f:\mathcal{C}\rightarrow \mathcal{Y}\) takes an input \(x\) and predicts its label \(f(x)\). Let \(S_m(x,y)\) denote the confidence value for the \(y^{th}\) category at the softmax layer for input \(x\). Let \(Syn(w, \delta, k)\) represent the set of the first \(k\) synonyms of \(w\) within  distance \(\delta\) in the embedding space, namely   
\begin{equation*}
\begin{gathered}
  Syn(w, \delta, k) = \{ \hat{w}^1, \dots, \hat{w}^i, \dots, \hat{w}^k | \hat{w}^i\in \mathcal{W}\\
    \ \wedge \|w-\hat{w}^1\|_p \leq ... \leq \|w-\hat{w}^k\|_p < \delta \},
\end{gathered}
\end{equation*}
where \(\|w-\hat{w}\|_p\) is the \(p\)-norm distance and we use Euclidean distance ($p=2$) in this work.

\subsection{Textual Adversarial Examples}
Suppose we have an oracle classifier \(\mathit{c}:\mathcal{C}\rightarrow \mathcal{Y}\) that could always output the correct label for any input text. For a subset (training set or test set) of texts \(\mathcal{T} \subseteq \mathcal{C}\) and a small constant \(\epsilon\), we could define the natural language adversarial examples as following:
\begin{align*}
        \mathcal{A} = \{&x_{adv}\in \mathcal{C} ~|~ \exists x \in \mathcal{T},d(x,x_{adv}) < \epsilon \ \wedge \\  
    &f(x_{adv}) \neq \mathit{c}(x_{adv}) = \mathit{c}(x) = f(x)  \},
\end{align*}
where \(d(x,x_{adv})\) 
is a distance metric that evaluates the dissimilarity between the benign example \(x=\langle w_1, \dots, w_i, \dots, w_n\rangle\) and the adversarial example \(x_{adv} = \langle w_1', \dots, w_i', \dots, w_n' \rangle\). In word-level attacks, \(d(\cdot, \cdot)\) is usually defined as the $p$-norm distance:
\begin{equation*}
    d(x,x_{adv})=\|x - x_{adv} \|_p = \left(\sum_i \|w_i - w'_i\|_p\right)^{\frac{1}{p}}.
\end{equation*}

\subsection{Textual Adversarial Attacks}\label{sec:bg:attack}
In recent years, various adversarial attacks for text classification have been proposed, including character-level, word-level and sentence-level attacks. 
\citet{ebrahimi2017hotflip} propose a method called HotFlip that swaps characters for character-level attack based on cost gradients. 
\citet{li2018textbugger} propose TextBugger that considers mostly character-level perturbations with some word-level perturbations by inserting, removing, swapping and substituting letters or replacing words. For a more combined approach, \citet{liang2017deep} propose to attack the target model by inserting Hot Training Phrases (HTPs) and modifying or removing Hot Sample Phrases (HSPs). Similarly, \citet{samanta2017crafting} propose to remove or replace important words or introduce new words in the text to craft adversarial examples. On the sentence level, \citet{iyyer2018adversarial} propose syntactically controlled paraphrase networks (SCPNs) to generate adversarial examples by rephrasing the sentence. Additionally, \citet{ribeiro2018semantically} generalize adversaries into semantically equivalent adversarial rules (SEARs).

Among all the types of adversarial attacks, synonyms substitution based attack \citep{kuleshov2018adversarial,alzantot2018generating,Shuhuai2019wordsaliency, yuan2020PSO, puyudi2020ws} is the representative method because it satisfies the lexical, grammatical and semantic constraints and is harder to be detected by both automatic and human investigation.
Here we provide a brief overview of three popular synonym substitution based adversarial attack methods.

\citet{kuleshov2018adversarial} propose a \textit{Greedy Search Algorithm (GSA)} that substitutes words with their synonyms so as to maintain the semantic and syntactic similarity. Specifically, given an input text $x$, GSA first constructs a synonym set $W_s$ for all words $w_i \in x$. Then at each step, GSA greedily chooses a word $\hat{w}_i' \in W_s$ that minimizes the confidence value $S_m(\hat{x}, y_{true})$, where $\hat{x}=\langle w'_1, \dots, w'_{i-1}, \hat{w}'_i,w'_{i+1}, \dots, w'_n\rangle$.

\citet{alzantot2018generating} propose a \textit{Genetic Algorithm (GA)} 
with two main operators:
1) \textit{Mutate(x)} randomly chooses a word $w_i \in x$ and replaces $w_i$ with \(\hat{w}_i\). Here, \(\hat{w}_i\) is determined as one of the synonyms \(Syn(w_i,\delta, k)\) that does not violate the syntax constraint imposed by the Google one billion words language model \citep{chelba2013billion} and minimizes the confidence value on category $y_{true}$. 2) \textit{Crossover($x_1, x_2$)} randomly chooses a word at each position from the candidate adversarial examples $x_1$ or $x_2$ to construct a new text $x$. They adopt these two operators to iteratively generate populations of candidate adversaries until there exists at least one successful adversarial example in the current population.

\citet{Shuhuai2019wordsaliency} propose a novel synonym substitution based attack method called \textit{Probability Weighted Word Saliency (PWWS)}, which considers the word saliency as well as the classification confidence. They define word saliency as the confidence change after removing this word temporarily. PWWS greedily substitutes word $w_i \in x$ with its optimal synonym $\hat{w}_i^*$, where $w_i$ has the maximum score on the combination of classification confidence change and word saliency among the unreplaced words.

\begin{figure*}[htbp]
    \centering
    \includegraphics[width=\textwidth]{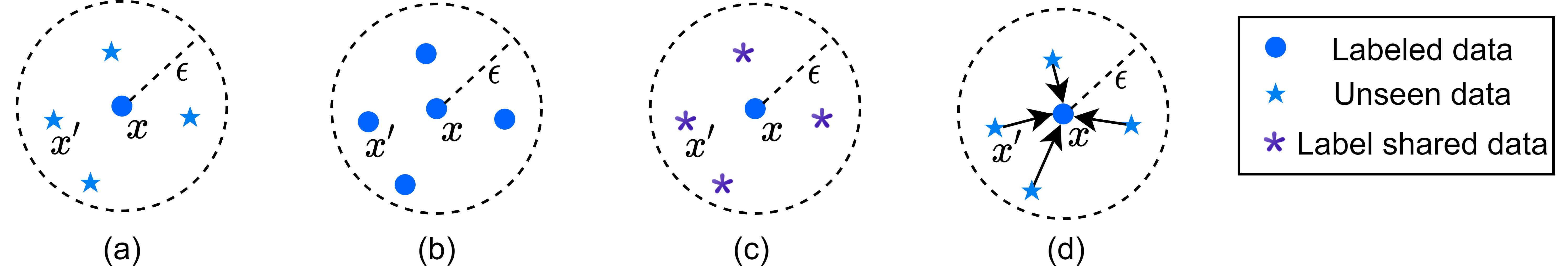}
    \caption{The neighborhood of a data point \(x\) in the input space. (a) Normal training: there exists some data point \(x'\) that the model has never seen before and yields wrong classification. 
    (b) Adding infinite labeled data: this is an ideal case that the model has seen all possible data points to resist adversaries. 
    (c) Sharing label: all the neighbors share the same label with \(x\).
    (d) Mapping neighborhood data points: mapping all neighbors to center \(x\) so as to eliminate adversarial examples.}
    \label{fig:adv_illustration}
\end{figure*}

\subsection{Textual Adversarial Defenses}
As text adversarial attacks have only attracted increasing interest since 2018, 
up to now there are relatively few works on adversarial defenses.

On the character-level, \citet{pruthi2019combating} propose to place a word recognition model in front of the downstream classifier to defend adversarial spelling mistakes. \citet{jones2020robust} propose Robust Encodings (RobEn) that maps the input sentences to a smaller, discrete space of encodings so as to eliminate various adversarial typos. \citet{hofmann2020superbizarre} propose Base-Inflection Encoding (BITE) that tokenizes English text by reducing inflected words to their base forms to generate robust symbol sequences against the inflectional adversarial examples. 

On the word-level, \citet{alzantot2018generating} and \citet{Shuhuai2019wordsaliency} incorporate their generated adversarial examples at the training stage to elevate the model robustness. Notice that \citet{iyyer2018adversarial} also include their generated adversarial paraphrases during the training to augment the training data. However, such data augmentation technique is subject to the limit of adversarial examples that could not be efficiently generated during training. 
To cover all possible word substitutions of an input, \citet{jia2019certified} and \citet{huang2019achieving} target certified robustness based on Interval Bound Propagation (IBP)~\citep{Gowal2019IBP} , i.e., to provide a provable guarantee that the model is robust to all word substitutions in this sample. Such defenses, however, are hard to scale to large datasets and neural networks such as BERT due to the high complexity, and result in lower accuracy on benign data due to the looser upper bounds. 

Different from adversarial training which incorporates extra adversarial examples and IBP which modifies the architecture, our work trains the model with an encoder for synonyms embedded in front of the input layer with normal training to improve the model robustness.

\section{Methodology}
In this section, we first introduce our motivation, then present the proposed \textit{Synonym Encoding Method} (SEM) for adversarial defense.

\subsection{Motivation}

Let \(\mathcal{X}\) denote the input space and \(V_\epsilon(x)\) denote the \(\epsilon\)-neighborhood of a data point \(x \in \mathcal{X}\), where \(V_\epsilon(x)=\{x'\in \mathcal{X} | \|x'-x\|_p < \epsilon\}\). As illustrated in Figure \ref{fig:adv_illustration} (a), we postulate that the weak generalization of the model leads to the existence of adversarial examples. Specifically, for any data point \(x \in \mathcal{X}\), \(\exists x' \in V_\epsilon(x), f(x') \neq y'_{true}\) and \(x'\) is an adversarial example of \(x\).

Ideally, to defend adversarial attacks, we need to train a classifier \(f\) that not only guarantees \(f(x)=y_{true}\), but also assures \(\forall x' \in V_\epsilon(x), f(x')=y'_{true}\). Thus, one of the most effective ways is to add more labeled data to improve the adversarial robustness \citep{Schmidt2018AdvRobust}. As illustrated in Figure \ref{fig:adv_illustration} (b), with infinite labeled data, we can train a model \(f: \forall x' \in V_\epsilon(x), f(x') = y'_{true}\) with high probability so that model \(f\) is robust enough to adversaries. Practically, however, labeling data is very expensive, and it is impossible to have even approximately infinite labeled data. 

Thus, as illustrated in Figure \ref{fig:adv_illustration} (c), \citet{Wong2018Provable} propose to construct a convex outer bound and guarantee that all data points in this bound share the same label. The goal is to train a model \(f:\forall x' \in V_\epsilon(x), f(x') = f(x) = y_{true}\). Specifically, they propose a linear-programming (LP) based upper bound on the robust loss by adopting a linear relaxation of the ReLU activation and minimize this upper bound during the training. Then, they bound the LP optimal value and calculate the element-wise bounds on the activation functions based on a backward pass through the network. Although their method does not need any extra data, it is hard to scale to realistically-sized networks due to the high calculation complexity. Similarly, we find that Interval Bound Propagation (IBP)~\citep{Gowal2019IBP} based methods, that can offer certified defense in the text domain, also face such challenge of high computational cost and lead to low classification accuracy on benign data.

In this work, as illustrated in Figure \ref{fig:adv_illustration} (d), we propose a novel way to find an encoder \(E:\mathcal{X} \rightarrow \mathcal{X}\) where \(\forall x' \in V_\epsilon(x), E(x')=x\). In this way, we force the classification boundary to be smoother without requiring any extra data to train the model or modifying the model's architecture. All we need to do is to insert the encoder before the input layer and train the model on the original training set. Now the problem turns into locating the neighbors of a data point \(x\). For image tasks, it is hard to find all images in the neighborhood of \(x\) in the input space, because the continuity results in infinite neighbors. For NLP tasks, however, utilizing the property that words in sentences are discrete tokens, we can easily find almost all synonymous neighbors of an input text. Based on this insight, we propose a new method called \textit{Synonym Encoding} to locate the neighbors of the input text \(x\).

\subsection{Synonym Encoding}
We assume that a smaller distance between two sentences in the embedding space indicates a closer meaning of the two sentences without considering the rephrased synonymous sentences. Therefore, we suppose that the neighbors of \(x\) are its synonymous sentences. In order to find these sentences, a reliable way is to substitute the words in the original sentence with their close synonyms. In this way, to construct an encoder \(E\) that encodes a set of synonyms to the same code, we cluster the synonyms in the embedding space and allocate a unique token for each cluster. 
The details of synonym encoding are shown in Algorithm \ref{alg:synonyms_encoding}. 

\begin{algorithm}[t]
	\renewcommand{\algorithmicrequire}{\textbf{Input:}}
	\renewcommand{\algorithmicensure}{\textbf{Output:}}
	\caption{\textit{Synonym Encoding Algorithm}}
	\label{alg:synonyms_encoding}
	\begin{algorithmic}[1]
		\REQUIRE \(\mathcal{W}\): dictionary of words\\
		          ~~~~~~~~\(n\): size of \(\mathcal{W}\) \\
		          ~~~~~~~~\(\delta\): distance for synonyms\\ 
		          ~~~~~~~~\(k\): number of synonyms for each word
		\ENSURE \(E\): encoding result
		
		\STATE \(E = \{w_1:\text{None}, \dots, w_n:\text{None}\}\)
		\STATE Sort the words dictionary \(\mathcal{W}\) by word frequency
		\FOR{each word \(w_i \in \mathcal{W}\)}
    		\IF{\(E[w_i] = \) NONE}
    		    \IF{\(\exists \hat{w}_i^j \in Syn(w_i, \delta, k)\), \(E[\hat{w}_i^j] \neq \) NONE}
        		    \STATE \(\hat{w}_i^*\leftarrow\) the closest encoded synonym \(\hat{w}_i^j \in Syn(w_i, \delta, k)\) to $w_i$
        		    \STATE \(E[w_i] = E[\hat{w}_i^*]\)
    		    \ELSE
    		        \STATE \(E[w_i]=w_i\)
    		    \ENDIF
    		    \FOR{each word \(\hat{w}_i^j\) in \(Syn(w_i, \delta, k)\)}
    		        \IF{\(E[\hat{w}_i^j]=\) NONE}
    		            \STATE \(E[\hat{w}_i^j] = E[w_i]\)
    		        \ENDIF
    		    \ENDFOR
    		\ENDIF
		\ENDFOR
		\STATE \textbf{return} \(E\)
	\end{algorithmic}  
\end{algorithm} 
Basically, we iterate through the word dictionary in the descending order of word frequency and try to find suitable code for each word. For a word $w_i$ that is not encoded, we find its synonym set by $Syn(w_i,\delta,k)$ and let its code be the encoding of its closest encoded synonym if there exists any, otherwise we set the code to be the word itself. We further propagate this code to any of its non-encoded synonyms. In this way, we obtain an encoder that automatically finds synonym clusters of various sizes and provides for the words in each cluster the same code with the highest frequency. Note that in our experiment, we implement the synonym encoding on GloVe vectors after counter-fitting \citep{mrkvsic2016counter}, which injects antonymy and synonymy constraints into the vector space representations so as to remove antonyms from being considered as similar words. Moreover, the hyper-parameter $k$, the number of synonyms we consider for each word, and $\delta$, the upper bound for the distance between the original word and its synonyms in the embedding space, are determined through experiments. A too small value of $k$ or $\delta$ would result in an insufficient cluster, while a too large value would cause the cluster to include words that are not close synonyms to each other. Through careful experimental study, we find $k = 10$ and $\delta = 0.5$ a proper choice with regard to the trade-off between generalization and robustness. 

After obtaining the encoder $E$, we can train the model with $E$ embedded before the input layer using normal training. Note that the encoder is only based on the given dictionary and dataset, and is unrelated to the model.

\section{Experiments}
To validate the efficacy of SEM, we take IBP and adversarial training as our baselines and evaluate the performance of SEM against three synonym substitution based attacks, namely GSA, PWWS and GA, on three popular benchmark datasets involving CNN, RNN and BERT models. 

\subsection{Experimental Setup}
We first provide an overview of datasets, classification models and baselines used in experiments.

\textbf{Datasets.}~~We select three popular datasets: \textit{IMDB}, \textit{AG's News}, and \textit{Yahoo! Answers}. \textit{IMDB} \citep{Potts2019IMDB} is a large dataset for binary sentiment classification, containing \(25,000\) highly polarized movie reviews for training and \(25,000\) for testing. \textit{AG's News} \citep{Zhang2015Zhang} consists of news articles pertaining four classes: World, Sports, Business and Sci/Tech. Each class contains \(30,000\) training examples and \(1,900\) testing examples. \textit{Yahoo! Answers} \citep{Zhang2015Zhang} is a topic classification dataset from the ``Yahoo! Answers Comprehensive Questions and Answers" version 1.0 dataset with 10 categories, such as Society \& Culture, etc. Each class contains 140,000 training samples and 5,000 testing samples.

\textbf{Models.}~~To evaluate the effectiveness of our method, we adopt several state-of-the-art models for text classification, including Convolution Neural Networks (CNNs), Recurrent Neural Networks (RNNs) and BERT. The embedding dimension for all CNN and RNN models are 300 \citep{Mikolov2013Efficient}. 
We replicate the CNN's architecture from \citet{kim2014convolutional}, that contains three convolutional layers with filter size of 3, 4, and 5 respectively, a max-pooling layer and a fully-connected layer. LSTM consists of three LSTM layers where each layer has \(128\) LSTM units and a fully-connected layer \citep{liu2016recurrent}. Bi-LSTM contains a bi-directional LSTM layer whose forward and reverse have \(128\) LSTM units respectively and a fully-connected layer. For the BERT model, we fine-tune base-uncased BERT \citep{devlin2018bert} using the corresponding dataset.

\textbf{Baselines.}~~We take adversarial training \citep{Goodfellow2015Explaining} and the certified defense IBP \citep{jia2019certified} as our baselines. We adopt three synonym substitution based attacks, GSA \citep{kuleshov2018adversarial}, PWWS \citep{Shuhuai2019wordsaliency} and GA \citep{alzantot2018generating} (described in Section \ref{sec:bg:attack}), to evaluate the defense performance of baselines and  SEM. However, due to the low efficiency of text adversarial attacks, we cannot implement adversarial training as it is in the image domain. In experiments, we adopt PWWS, which is faster than GA and more effective than GSA, to generate \(10\%\) adversarial examples of the training set, and re-train the model incorporating adversarial examples with the training data. Besides,
as \cite{zhou20robustness} point out that large-scale pre-trained models such as BERT are too challenging to be tightly verified with current technologies by IBP, we do not adopt IBP as the baseline on BERT. For fair comparison, we construct the synonym set using GloVe vectors after counter-fitting for all methods.

\begin{table*}[tb]
\centering
\caption{The classification accuracy (\(\%\)) of various models on three datasets, with or without defense methods, on benign data or under adversarial attacks. For each model (Word-CNN, LSTM, Bi-LSTM or BERT), the highest classification accuracy for various defense methods is highlighted in \textbf{bold} to indicate the \textbf{best defense efficacy}. NT: Normal Training, AT: Adversarial Training.}
\scalebox{0.88}{
\begin{tabular}{cccccc|cccc|cccc|ccc}
\toprule
\multirow{2}{*}{Dataset}& \multirow{2}{*}{Attack} & \multicolumn{4}{c}{Word-CNN} & \multicolumn{4}{c}{LSTM} & \multicolumn{4}{c}{Bi-LSTM} & \multicolumn{3}{c}{BERT}\\
\cmidrule(r){3-17} 
 & & NT & AT & IBP & SEM & NT & AT & IBP & SEM & NT & AT & IBP & SEM & NT & AT & SEM \\
 \midrule
\multirow{4}{*}{\textit{IMDB}}
& No-attack & 88.7    & \textbf{89.1} & 78.6 & 86.8 & 87.3   & \textbf{89.6}  & 79.5 & 86.8 & 88.2   & \textbf{90.3}  & 78.2 & 87.6 & 92.3 & \textbf{92.5} & 89.5\\
& GSA   & 13.3    & 16.9  & \textbf{72.5} & 66.4 & ~~8.3  & 21.1 & 70.0 & \textbf{72.2} &  ~~7.9 & 20.8  & \textbf{74.5} & 73.1 & 24.5 & 34.4 & \textbf{89.3} \\
& PWWS  & ~~4.4   & ~~5.3 & \textbf{72.5} & 71.1 & ~~2.2  & ~~3.6 & 70.0 & \textbf{77.3} &  ~~1.8 & ~~3.2 & 74.0 & \textbf{76.1} & 40.7 & 52.2 & \textbf{89.3} \\
& GA    & ~~7.1   & 10.7  & 71.5 & \textbf{71.8} & ~~2.6  & ~~9.0 & 69.0 & \textbf{77.0} &  ~~1.8 & ~~7.2 & \textbf{72.5} & 71.6 & 40.7 & 57.4 & \textbf{89.3} \\
\midrule
\multirow{4}{*}{\shortstack{\textit{AG's} \\ \textit{News}}}
& No-attack & \textbf{92.3} & 92.2 & 89.4 & 89.7 & 92.6 & \textbf{92.8} & 86.3 & 90.9 & \textbf{92.5} & \textbf{92.5} & 89.1 & 91.4 & 94.6 & \textbf{94.7} & 94.1 \\
& GSA   & 45.5 & 55.5 & \textbf{86.0} & 80.0 & 35.0 & 58.5 & 79.5 & \textbf{85.5} & 40.0 & 55.5 & 79.0 & \textbf{87.5} & 66.5 & 74.0 & \textbf{88.5} \\
& PWWS  & 37.5 & 52.0 & \textbf{86.0} & 80.5 & 30.0 & 56.0 & 79.5 & \textbf{86.5} & 29.0 & 53.5 & 75.5 & \textbf{87.5} & 68.0 & 78.0 & \textbf{88.5} \\
& GA    & 36.0 & 48.0 & \textbf{85.0} & 80.5 & 29.0 & 54.0 & 76.5 & \textbf{85.0} & 30.5 & 49.5 & 78.0 & \textbf{87.0} & 58.5 & 71.5 & \textbf{88.5}\\
\midrule
\multirow{4}{*}{\shortstack{\textit{Yahoo!} \\\textit{Answers}}}
& No-attack & 68.4  & \textbf{69.3} & 64.2 & 65.8 & 71.6 & \textbf{71.7} & 51.2 & 69.0 & 72.3 & \textbf{72.8} & 59.0 & 70.2 & \textbf{77.7} & 76.5 & 76.2 \\
& GSA   & 19.6  & 20.8 & \textbf{61.0} & 49.4 & 27.6 & 30.5 & 30.0 & \textbf{48.6} & 24.6 & 30.9 & 39.5 & \textbf{53.4} & 31.3 & 41.8 & \textbf{66.8} \\
& PWWS  & 10.3  & 12.5 & \textbf{61.0} & 52.6 & 21.1 & 22.9 & 30.0 & \textbf{54.9} & 17.3 & 20.0 & 40.0 & \textbf{57.2} & 34.3 & 47.5 & \textbf{66.8}\\
& GA    & 13.7  & 16.6 & \textbf{61.0} & 59.2 & 15.8 & 17.9 & 30.5 & \textbf{66.2} & 13.0 & 16.0 & 38.5 & \textbf{63.2} & 15.7 & 33.5 & \textbf{66.4}\\
\bottomrule
\end{tabular}
}
\label{tab:Attack}
\end{table*}

\begin{table*}[tb]
\centering
\caption{The classification accuracy (\(\%\)) of various models for adversarial examples generated through other models on \textit{AG's News} for evaluating the transferability. * indicates that the adversarial examples are generated based on this model.}
\scalebox{0.9}{
\begin{tabular}{ccccc|cccc|cccc|ccc}
\toprule
\multirow{2}{*}{Attack}  & \multicolumn{4}{c}{Word-CNN} & \multicolumn{4}{c}{LSTM} & \multicolumn{4}{c}{Bi-LSTM} & \multicolumn{3}{c}{BERT}\\
\cmidrule(r){2-16} 
  & NT & AT & IBP & SEM & NT & AT & IBP & SEM & NT & AT & IBP & SEM & NT & AT & SEM\\
\midrule
GSA   & ~~45.5*   & 86.0  & \textbf{87.0} & \textbf{87.0} & 80.0  & 89.0  & 83.0 & \textbf{90.5} & 80.0 & 87.0 & 87.5 & \textbf{91.0} & 92.5 & \textbf{94.5} & 90.5 \\
PWWS  & ~~37.5*   & 86.5 & \textbf{87.0} & \textbf{87.0} & 70.5  & 87.5 & 83.0 & \textbf{90.5} &  70.0 & 87.0 & 86.5 & \textbf{90.5} & 90.5 & \textbf{95.0} & 90.5 \\
GA    & ~~36.0*   & 85.5  & \textbf{87.0} & \textbf{87.0} & 75.5 & 88.0 & 83.5 & \textbf{90.5} &  76.0 & 86.5 & 86.0 & \textbf{91.0} & 91.5 & \textbf{95.0} & 90.5 \\
\midrule
GSA   & 84.5 & 89.0 & \textbf{87.5} & 87.0 & ~~35.0* & 87.0 & 83.5 & \textbf{90.5} & 73.0 & 85.0 & 86.5 & \textbf{91.0} & 93.0 & \textbf{95.5} & 90.5 \\
PWWS  & 83.0 & 89.0 & \textbf{87.5} & 87.0 & ~~30.0* & 86.0 & 85.0 & \textbf{90.5} & 67.5 & 85.5 & 86.5 & \textbf{90.5} & 93.0 & \textbf{95.0} & 90.5 \\
GA    & 84.0 & 89.5 & \textbf{87.5} & 87.0 & ~~29.0* & 88.0 & 83.5 & \textbf{90.5} & 70.5 & 87.5 & 87.0 & \textbf{91.0} & 92.5 & \textbf{95.5} & 90.5 \\
\midrule
GSA   & 81.5  & \textbf{88.0} & 87.5 & 87.0 & 72.5 & 89.5 & 84.0 & \textbf{90.5} & ~~40.0* & 85.5 & 87.5 & \textbf{91.0} & 93.5 & \textbf{95.5} & 91.0 \\
PWWS  & 80.0  & \textbf{87.0} & \textbf{87.0} & 86.5 & 67.5 & 87.5 & 83.5 & \textbf{90.5} & ~~29.0* & 85.5 & 87.0 & \textbf{90.5} & 92.5 & \textbf{95.5} & 90.5 \\
GA    & 80.0  & \textbf{89.5} & 87.5 & 87.0 & 69.5 & 88.5 & 83.5 & \textbf{90.5} & ~~30.5* & 85.0 & 86.5 & \textbf{90.5} & 92.5 & \textbf{95.0} & 90.5 \\
\midrule
GSA & 83.5 & 87.0 & \textbf{87.5} & 87.0 & 84.0 & 88.0 & 83.5 & \textbf{89.5} & 83.0 & 88.0 & 87.0 & \textbf{89.5} & ~~66.5* & \textbf{95.5} & 90.5 \\
PWWS & 81.0 & 87.5 & \textbf{88.0} & 87.0 & 82.5 & 88.0 & 84.0 & \textbf{91.5} & 83.0 & 88.0 & 87.5 & \textbf{91.5} & ~~68.0* & \textbf{94.5} & 90.5 \\
GA & 82.0 & 87.0 & \textbf{88.0} & 87.0 & 82.0 & 88.0 & 83.5 & \textbf{91.0} & 82.0 & 88.0 & 87.5 & \textbf{91.0} & ~~58.5* & \textbf{94.0} & 90.0 \\
\bottomrule
\end{tabular}
}
\label{tab:Transferability}
\end{table*}
\subsection{Evaluation on Defense Efficacy}
To evaluate the efficacy of SEM, we randomly sample \(200\) correctly classified examples on different models from each dataset and use the above adversarial attacks to generate adversarial examples on the target models with or without defense. The more effective the defense method is, the less the classification accuracy the model drops. Table \ref{tab:Attack} demonstrates the performance of various defense methods on benign examples or under adversarial attacks.

We could check each row to find the best defense results for each model under the setting of no-attack, GSA, PWWS, and GA attacks: 

\begin{itemize}[leftmargin=14pt,topsep=0.5pt, itemsep=1pt]
\item Under the setting of no-attack, adversarial training (AT) could improve the classification accuracy of most models on three datasets, as adversarial training (AT) also augments the training data. However, IBP achieves much lower accuracy on benign data due to its high complexity and looser upper bounds. Our defense method SEM reaches an accuracy that is very close to the normal training (NT), with a small trade-off between robustness and accuracy. Such trade-off is also common for defense methods in the image domain that has been theoretically studied \citep{Hongyang2019TRADE, Dimitris2019Robustness}. As discussed in Section \ref{sec:dis}, we select suitable hyper-parameters according to this trade-off for the best joint performance. 

\item Under the three different attacks, however, both the classification accuracy with normal training (NT) and adversarial training (AT) drop significantly. For normal training (NT), the accuracy degrades more than \(51\%\), \(26\%\) and \(43\%\) on the three datasets,  respectively. And adversarial training (AT) cannot defend these attacks effectively, especially for PWWS and GA on \textit{IMDB} and \textit{Yahoo! Answers} with CNN and RNN models, where adversarial training (AT) only improves the accuracy by a small amount (smaller than \(5\%\)). One possible reason is that adversarial training (AT) needs massive adversarial examples, which are much more than the benign examples, to improve the robustness, but adversarial training (AT) here in the text domain could not obtain enough adversarial examples on the current model due to the low efficiency of existing adversary generation. In contrast, SEM can remarkably improve the robustness of the deep learning models under all the three attacks and achieve the best robustness on LSTM, Bi-LSTM and BERT models on the three datasets. Note that IBP, firstly proposed for images, is more suitable for CNN models but does not perform very well on RNN models. Moreover, on the more complex dataset \textit{Yahoo! Answers}, SEM converges more quickly than normal training due to the simplicity of encoded space, while IBP is very hard to train and cannot achieve good performance on either benign data or adversarial examples due to its high complexity for training.
\end{itemize}

Furthermore, there might be a concern that mapping all synonyms into a unique encoding could harm the subtle linguistic distinctions or even cause that the words in the same cluster would not always be synonyms in different contexts. To explore whether this concern matters, we feed perturbed texts, which are generated by randomly picking 10\% words in the testing samples of \textit{AG's News} dataset and substituting them with arbitrary words in the dictionary, to normally trained (NT) CNN. We find that the model accuracy only decays by 2.4\%, indicating that deep neural models for text classification are robust to such interference. As previously mentioned, SEM exhibits a little decay on the classification accuracy of benign data, which is also consistent with the result of random substitution test. Thus, such concern does not significantly affect the robustness and stability of SEM.

\begin{figure*}[tb]
    \centering
    \includegraphics[width=\textwidth]{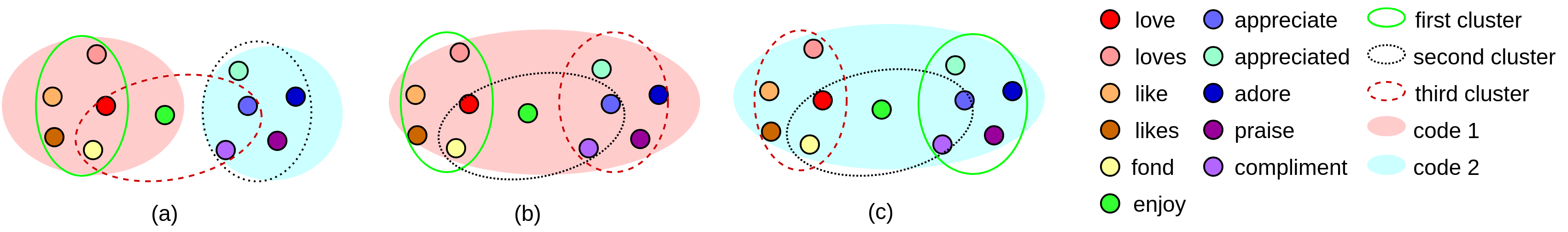}
    \caption{An illustration for various orders to traverse words at the 3rd line of Algorithm \ref{alg:synonyms_encoding} in the embedding space. 
    (a) Traverse words first on the left, then on the right, then in the middle. The synonyms are encoded into two various codes (left and right).
    (b) Traverse words first on the left, then in the middle, then on the right. All synonyms are encoded into a unique code of the left.
    (c) Traverse words first on the right, then in the middle, then on the left. All synonyms are encoded into a unique code of the right.}
    \label{fig:order}
\end{figure*}

\begin{figure}[ht]
    \centering
    \includegraphics[width=1.1\linewidth]{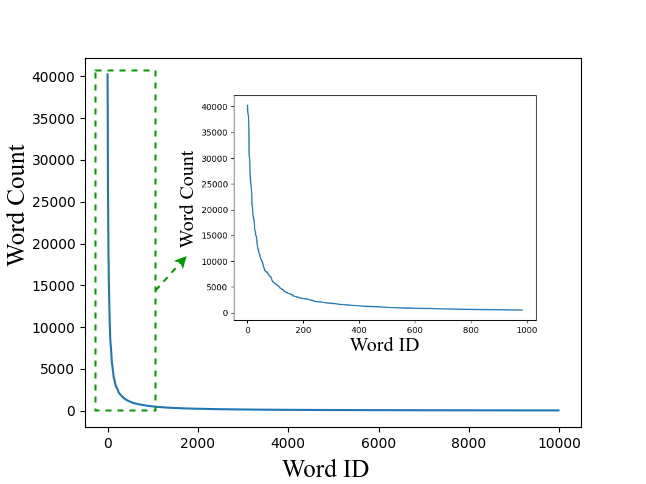}
    \caption{Word frequency of each word in \textit{IMDB} dataset.}
    \label{fig:wordFrequency}
\end{figure}

\begin{figure*}[htbp]
    \centering
    \begin{subfigure}{.23\textwidth} 
      \centering 
      \includegraphics[width=1.60in]{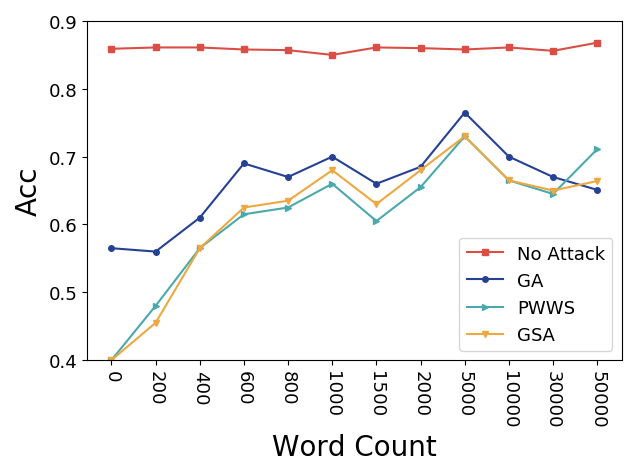}
      \caption{Word-CNN under attacks}
      \label{fig:orderAttack:first}
    \end{subfigure} 
    \quad
    \begin{subfigure}{.23\textwidth}
      \centering 
      \includegraphics[width=1.60in]{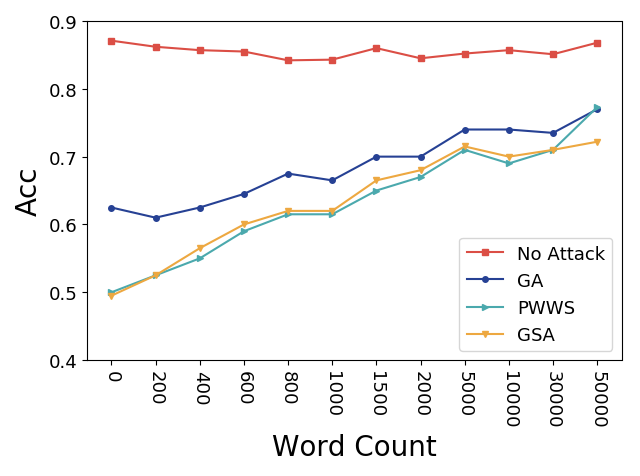}
      \caption{LSTM under attacks}
      \label{fig:orderAttack:second}
    \end{subfigure}
    \quad
    \begin{subfigure}{.23\textwidth}
      \centering 
      \includegraphics[width=1.60in]{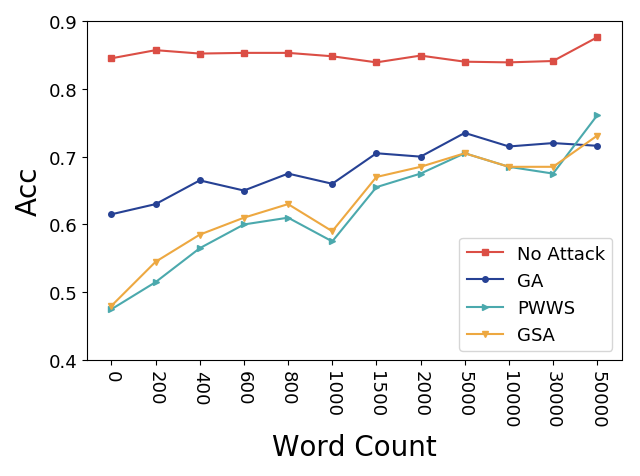}
      \caption{Bi-LSTM under attacks}
      \label{fig:orderAttack:third}
    \end{subfigure} 
    \quad
    \begin{subfigure}{.23\textwidth}
      \centering 
      \includegraphics[width=1.60in]{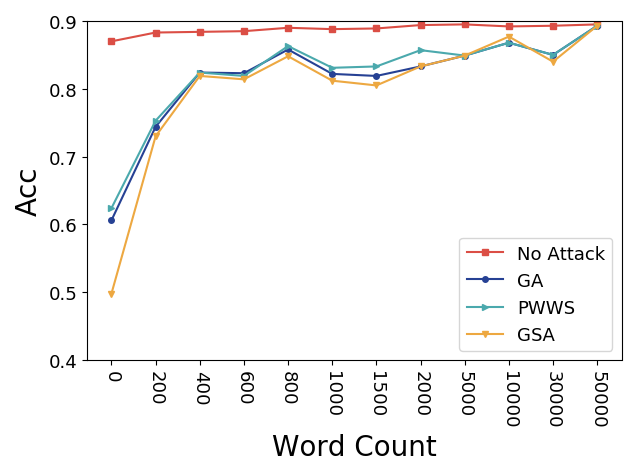}
      \caption{BERT under attacks}
      \label{fig:orderAttack:forth}
    \end{subfigure} 
    \caption{The impact of word frequency on the performance of SEM for four models on \textit{IMDB}. We report the classification accuracy (\%) of each model with various number of words ordered by word frequency.}
    \label{fig:orderAttack}
\end{figure*}
\begin{figure*}[htbp]
    \centering
    \begin{subfigure}{.23\textwidth}
      \centering 
      \includegraphics[width=1.60in]{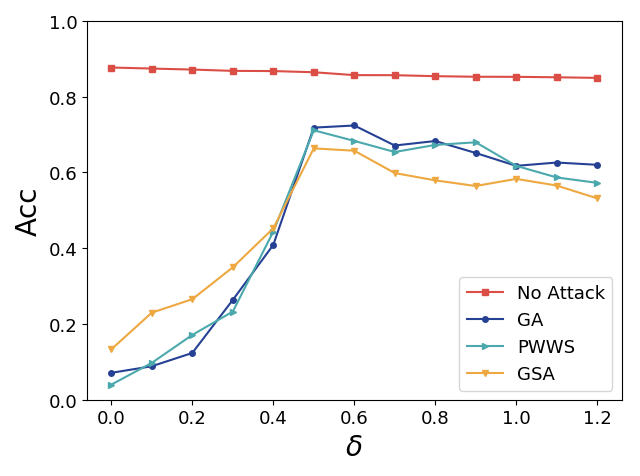}
      \caption{Word-CNN under attacks}
      \label{fig:ablation:first}
    \end{subfigure}
    \quad
    \begin{subfigure}{.23\textwidth} 
      \centering 
      \includegraphics[width=1.60in]{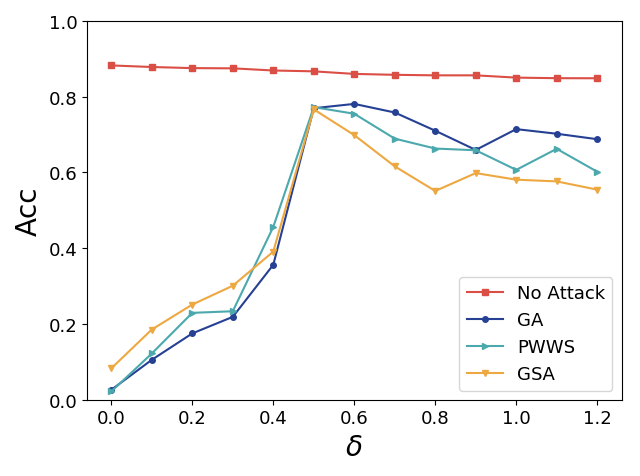}
      \caption{LSTM under attacks}
      \label{fig:ablation:second}
    \end{subfigure}%
    \quad
    \begin{subfigure}{.23\textwidth}
      \centering 
      \includegraphics[width=1.60in]{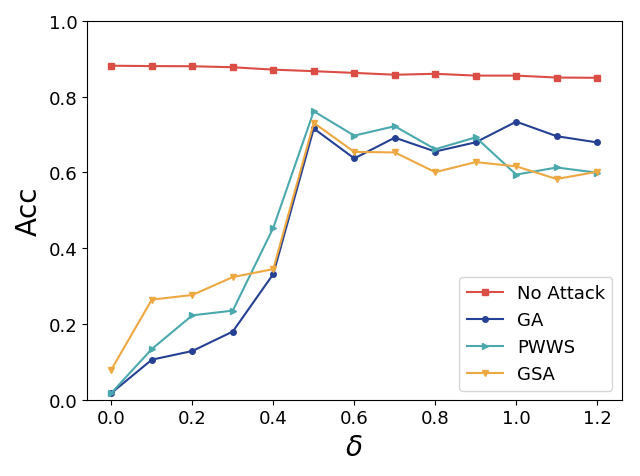}
      \caption{Bi-LSTM under attacks}
      \label{fig:ablation:third}
    \end{subfigure}
    \quad
    \begin{subfigure}{.23\textwidth}
      \centering 
      \includegraphics[width=1.60in]{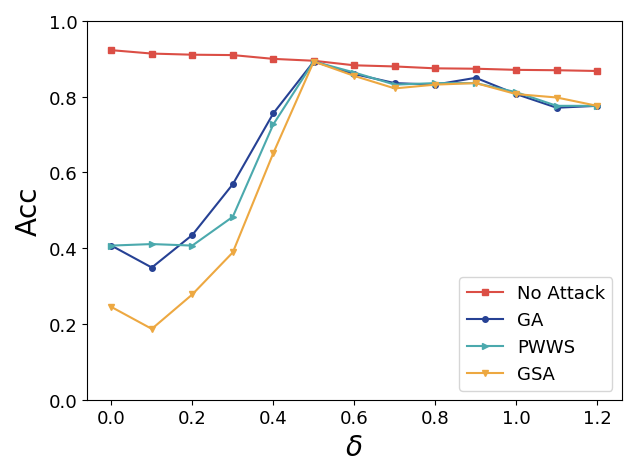}
      \caption{BERT under attacks}
      \label{fig:ablation:fourth}
    \end{subfigure}
    \caption{Classification accuracy (\%) of SEM on various values of \(\delta\) ranging from \(0\) to \(1.2\) for four models on \textit{IMDB} where \(k\) is fixed to \(10\).}
    \label{fig:ablation}
\end{figure*}

\begin{figure*}[!h]
    \centering
    \begin{subfigure}{.23\textwidth}
      \centering 
      \includegraphics[width=1.60in]{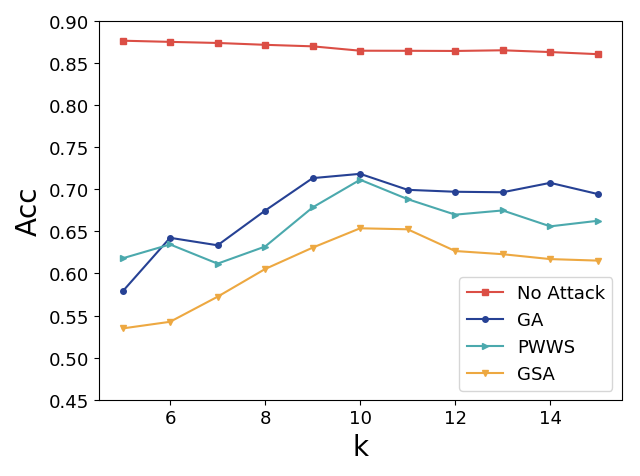}

      \caption{Word-CNN under attacks}
      \label{fig:ablation_k:first}
    \end{subfigure}%
    \quad
    \begin{subfigure}{.23\textwidth} 
      \centering 
      \includegraphics[width=1.60in]{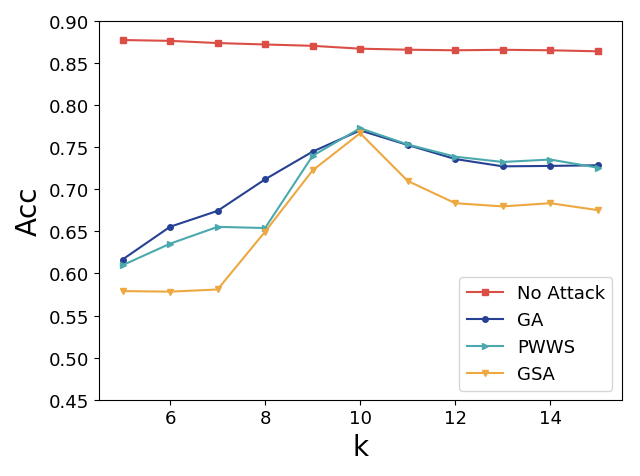}
      \caption{LSTM under attacks}
      \label{fig:ablation_k:second}
    \end{subfigure}%
    \quad
    \begin{subfigure}{.23\textwidth}
      \centering 
      \includegraphics[width=1.60in]{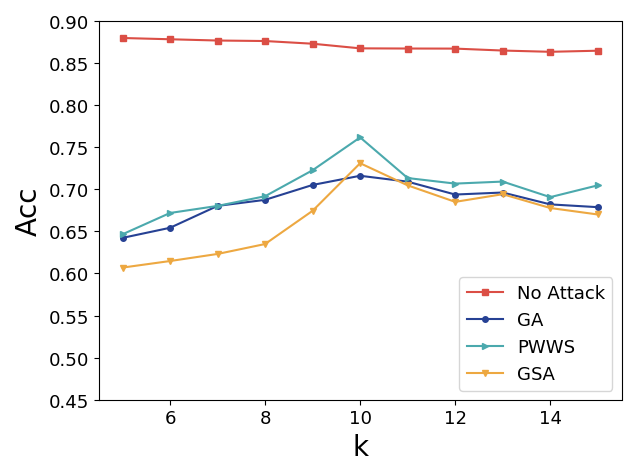}
      \caption{Bi-LSTM under attacks}
      \label{fig:ablation_k:third}
    \end{subfigure}
    \quad
    \begin{subfigure}{.23\textwidth}
      \centering 
      \includegraphics[width=1.60in]{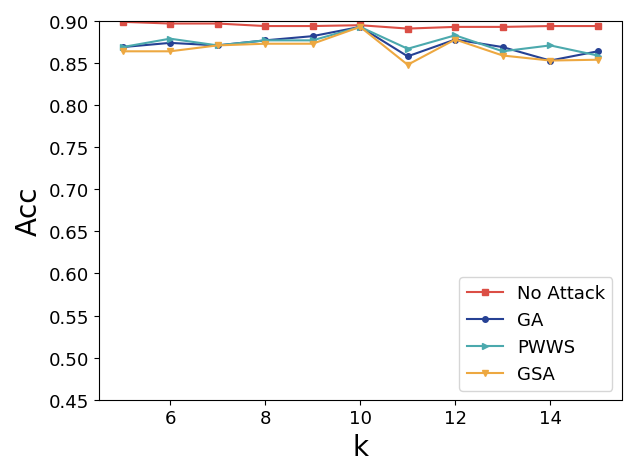}
      \caption{BERT under attacks}
      \label{fig:ablation_k:fourth}
    \end{subfigure} 
    \caption{Classification accuracy (\%) of SEM on various values of \(k\) ranging from \(5\) to \(15\) for four models on \textit{IMDB} where \(\delta\) is fixed to \(0.5\).}
    \label{fig:ablation_k}
\end{figure*}


\subsection{Defense against Transferability}
In the image domain, the transferability of adversarial attack refers to its ability to decrease the accuracy of different models using adversarial examples generated based on a specific model \citep{Goodfellow2015Explaining,Dong2018MIM,wang2021enhancing}, which is a more realistic threat. 
Therefore, a good defense method should not only defend the adversarial attacks but also resist the transferability of adversarial examples. 

To evaluate the ability of blocking the attack transferability, we generate adversarial examples on each model under normal training, and then test on other models with or without defense on \textit{AG's News} dataset. As shown in Table \ref{tab:Transferability}, almost on all RNN models with adversarial examples generated on other model, SEM could yield the highest classification accuracy. And on CNN models, SEM can achieve moderate accuracy on par with the best one. On BERT, the transferability of adversarial examples generated on other models performs very weak, and the accuracy here lies more on the generalization, so AT achieves the best results.

\subsection{Discussion on Traverse Order}
\label{sec:dis}
We further discuss the impact of the traverse order of synonymous words. 
As shown in Figure \ref{fig:order}, the traverse order of words at the 3rd line of Algorithm \ref{alg:synonyms_encoding} can influence the final synonym encoding of a word and even lead to different codes for the same synonyms set. In SEM, we traverse the word in the descending order of word frequency to allocate the encoding with the highest frequency to each word cluster. Hence, the encoded text tends to adopt codes of the more common words that are close synonyms to their original ones. 
The word frequency of each word in \textit{IMDB} is shown in Figure \ref{fig:wordFrequency}. 

To verify whether the order determined by word frequency could help SEM achieve higher robustness, we first traverse fixed number of words with the highest frequency (we choose $0$, $200$, $400$, $600$, $800$, $1,000$, $1,500$, $2,000$, $5,000$, $10,000$, $30,000$, $50,000$ respectively) and traverse the remaining words in arbitrary order to obtain a new encoder. The accuracy on benign data and robustness under attacks with different encoder on the four models are shown in Figure \ref{fig:orderAttack:first}-\ref{fig:orderAttack:forth}. As we can see, different traverse orders have little effect on the accuracy of benign data but indeed influence the robustness performance of SEM. On the four models, when we shuffle the entire dictionary for random traverse order (word count $= 0$), SEM achieves poor robustness but is still better than normal training and adversarial training. As we increase the number of fixed ordered words by word frequency, the robustness increases rapidly. When the word count is 5,000 for CNN and RNN models and 400 for BERT, SEM can achieve good enough robustness, and the best result on CNN models is even better than that of IBP. When we completely traverse the word dictionary according to the word frequency, SEM can achieve the best robustness on LSTM, Bi-LSTM and BERT. Therefore, the word frequency indeed has an impact on the performance of SEM. The higher frequency the word has, the more significant impact it has on the performance.

In summary, different orders to traverse words can influence the resulting encoding, and the order by word frequency can help improve the stability and robustness of SEM. 

\subsection{Hyper-Parameters Study}
\label{sec:param}
Moreover, we explore how the hyper-parameters \(\delta\) and \(k\) in \(Syn(w, \delta, k)\) of SEM influence its performance, using four models on \textit{IMDB} with or without adversarial attacks. We try different \(\delta\) ranging from \(0\) to \(1.2\) and \(k\) ranging from \(5\) to \(15\). The results are illustrated in Figure \ref{fig:ablation} and \ref{fig:ablation_k} respectively.


On benign data, as the red lines shown in Figure \ref{fig:ablation} and \ref{fig:ablation_k}, the classification accuracy decreases slightly when \(\delta\) or \(k\) increases, because a larger \(\delta\) or \(k\) indicates that we need fewer words to train the model. 
Nevertheless, the classification accuracy only degrades slightly, as SEM could maintain the semantic invariance of the original text after encoding.

Then, we investigate how \(\delta\), the distance we use to consider synonyms for a word, influences the defense performance of SEM empirically on the four models, as shown in Figure \ref{fig:ablation:first}-\ref{fig:ablation:fourth} where \(k\) is fixed to \(10\). 
When \(\delta = 0\), we have the original models, and the accuracy is the lowest under all attacks except for GA and GSA on BERT which achieve the lowest when \(\delta = 0.1\). 
As \(\delta\) increases, the accuracy rises rapidly, peaks when \(\delta = 0.5\), and then starts to decay because too large \(\delta\) introduces semantic drifts. Thus, we choose \(\delta = 0.5\) for a proper trade-off to maintain the accuracy of benign data and improve the robustness against adversarial examples.

Similarly, we investigate the influence of \(k\), the number of synonyms that we consider for each word, on the defense effectiveness of SEM on the four models, as shown in Figure \ref{fig:ablation_k:first}-\ref{fig:ablation_k:fourth} where \(\delta\) is fixed to \(0.5\). 
For BERT, \(k\) has little impact on the performance of SEM that could always effectively defend the attacks.
For CNN and RNN models, when \(k = 5\), some close synonyms cannot be encoded into the same code. However, we still observe that SEM improves the accuracy better than that of adversarial training obtained in previous experiments. As \(k\) increases, more synonyms are encoded into the same code, and thus SEM could defend the attacks more effectively. After peaking when \(k = 10\), the classification accuracy decays slowly and becomes stable if we continue to increase \(k\). 
Thus, we set \(k=10\) to achieve the trade-off on the classification accuracy of benign examples and adversarial examples.

In summary, small \(\delta\) or \(k\) results in some synonyms not being encoded correctly and leads to weak defense performance, while large \(\delta\) or \(k\) might cause SEM to cluster words that are not synonyms and degrade the defense performance. 
Therefore, we choose \(\delta = 0.5\) and \(k=10\) to have a good trade-off. 

\section{Conclusion}
In this work, we propose a new word-level adversarial defense method called \textit{Synonym Encoding Method} (SEM) for the text classification task in NLP. 
SEM encodes the synonyms of each word and embeds the encoder in front of the input layer of the neural network model. 
Compared with existing adversarial defense methods, adversarial training and IBP, SEM can effectively defend synonym substitution based attacks and block the transferability of adversarial examples, while maintaining good classification accuracy on the benign data. Besides, SEM is efficient and easy to scale to large models and big datasets. Further discussions are also provided on the traverse order of the synonym words, and the impact of hyper-parameters of SEM.

We observe that SEM not only promotes the model robustness, but also accelerates the training process due to the simplicity of encoding space. Considering the semantic consistency after replacing the words with synonyms, SEM has the potential to be adopted to other NLP tasks for adversarial defense, as well as for simplifying the training process. 



\begin{acknowledgements} 
    This work is supported by National Natural Science Foundation (62076105) and Microsoft Research Asia Collaborative Research Fund (99245180).

\end{acknowledgements}

\bibliography{wang_315}
\end{document}